\documentclass[letterpaper, 10 pt, conference]{ieeeconf}  

\usepackage{xcolor}

\usepackage{url}
\usepackage{comment}

\IEEEoverridecommandlockouts                              

\usepackage{amsmath} 
\newtheorem{definition}{Definition}
\usepackage{mathptmx}
\usepackage{amsfonts}
\usepackage{algorithm}
\usepackage{algorithmic}

\usepackage{tikz}
\usetikzlibrary{automata,positioning,arrows}

\newcommand\vu{\mathbf{u}}
\newcommand\vs{\mathbf{s}}
\newcommand{\F}{\Diamond}
\newcommand{\G}{\Box}
\newcommand{\U}{\mathcal{U}}
\newcommand{\spec}{\varphi}
\newcommand{\true}{\top}

\newcommand{\outsig}{y}

\newcommand{\Props}{P}
\newcommand{\RM}{\mathsf{R}}
\newcommand{\traj}{\xi}
\newcommand{\Traj}{\Xi}
\newcommand{\rewardStep}{\mathcal{R}}

\newcommand{\reals}{\mathbb{R}}
\newcommand{\minrho}{\underline{\rho}}
\newcommand{\maxrho}{\overline{\rho}}

\newcommand{\rlrom}{RLRom}
\newcommand{\RMSTL}{RM-STL}

\usepackage{amssymb,bm}
\usepackage{esvect}
\usepackage{siunitx}
\usepackage{subcaption}
\usepackage{balance}

\title{\LARGE \bf
On Tackling Complex Tasks with Reward Machines and Signal Temporal Logics
}

\author{Ana Mar\'ia G\'omez Ruiz$^{1}$, Thao Dang$^{2}$ and Alexandre Donz\'e$^{1}$
\thanks{$^{1}$VERIMAG, Universit\'e Grenoble Alpes
        {\tt\small {\{firstname.lastname\}@univ-grenoble-alpes.fr}}}%
\thanks{$^{2}$CNRS, VERIMAG, Universit\'e Grenoble Alpes
        {\tt\small {\{firstname.lastname\}@univ-grenoble-alpes.fr}}}%
}

\begin{document}

\begin{titlepage}

\begin{figure}
\includegraphics[height=2cm]{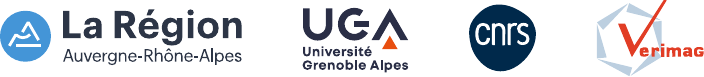} 
\vspace*{1cm}
\end{figure}

\maketitle

\end{titlepage}

\thispagestyle{empty}
\pagestyle{empty}

\begin{abstract}
  We propose a Reinforcement Learning (RL) based control design framework for handling complex tasks. The approach extends the concept of Reward Machines (RM) with Signal Temporal Logic (STL) formulas that can be used for event generation. The use of STL allows not only a more efficient representation of rewards for complex tasks but also guiding the training process to converge towards behaviors satisfying specified requirements. We also propose an implementation of the framework that leverages the STL online monitoring algorithms. We illustrate the framework with three case studies (minigrid, cart-pole and high-way environments) with non-trivial tasks. 
\end{abstract}

\section{INTRODUCTION}

RL has recently been proven to be efficient for solving decision-making problems in various domains, including classical control tasks, games, and multi-agent systems. A key element in RL is specifying a reward function that encodes the goal. However, this concept of reward is difficult to extend to complex task specifications, for example, a robot should reach a number of target points in a given order without entering some unsafe area. Such history-dependent specifications can be, on the other hand, described conveniently using temporal logic.   

Recently there has been significant research interest on using temporal logics for specifying RL tasks~\cite{BeltaCDC2016,Belta2017,Brafman2018,Hasanbeig2019,DeGiacomo2019,Hahn2019,Hasanbeig2020,Topcu2019,Kapoor2020,Kuo2020,Topcu2021,Vaezipoor2021,Jothimurugan2022}. Most of this work focuses on transforming specifications into a specification automaton, which is then composed with the Markov Decision Process (MDP) representing the environment to be controlled. Then the rewards are defined for the transitions of the resulting MDP, using reachability probabilities, to capture the acceptance conditions of the specification automaton. The theoretical questions on the optimality preservation of the transformation from Linear Temporal Logic (LTL) specifications to reward-based specifications are addressed in~\cite{Alur2022}.
A more direct approach~\cite{Icarte2018} is to define an automata-based model called RM to provide a modular and hierarchical framework for defining rewards in complex tasks. This approach can handle non-Markovian reward functions that depend not only on the current state but also on the evolution history of the environment. The authors also propose a Q-learning based algorithm, called QRM, to learn control policies. A later paper~\cite{Icarte2022} proposes variants of QRM including a hierarchical RL algorithm (HRM) to learn policies for tasks specified using RM.  In~\cite{Icarte2019} the authors demonstrate how to transform LTL specification into RM, while ~\cite{Topcu2021rm} proposes a notion of RM decomposition to achieve decentralized learning in a cooperative multi-agent context. Moving to continuous-time settings, in~\cite{MunirajCDC2018}, a multi-agent deep Q-learning algorithm is proposed to learn the optimal policies (in a finite-state multi-agent stochastic game framework) that maximize the expected robustness degree of the Signal Temporal Logic (STL) specification describing mission objectives, against the best responses of the adversarial agents.
To handle non-Markovian STL specifications,~\cite{Ikemoto2022} proposes $\tau$-dMDP, an extended MDP using past system states and control actions. Using the $\tau$-dMDP, a Deep RL algorithm is proposed to design a networked controller subject to network delays to complete an STL specification.  
In~\cite{BeltaCDC2016}, two problems of learning optimal policies for STL specifications are formulated as maximizing respectively the probability of satisfaction and the expected robustness degree. On this field of formal specification guided RL, the reader is also referred to a recent survey~\cite{Bansal2022}. 

While the non-Markovian reward issue is resolved, defining rewards to reflect complex control objectives and to enable efficient policy learning is still a challenge. Indeed, conventional learning algorithms often become trapped in local optima, or, when reward information is too sparse, may spend many iterations exploring experiences that do not meaningfully contribute to optimizing the reward. The goal of this paper is to address these issues by proposing an extension of the RM concept using STL. The motivation for this extension is twofold. First, STL being defined over real-valued signals, enables a more expressive and compact representation of history-dependent reward mechanisms. It also offers a principled way to characterize behaviors that are trapped near local optima, allowing the training process to identify and circumvent such suboptimal convergence. Second, while the concepts of RM and TL in discrete time have been extensively studied —with many theoretical aspects such as convergence and optimality already addressed— most existing methods rely on transforming TL into RM. This transformation typically produces large automata, which can obscure the interpretability of the original task specification. We propose a framework that combines RM with STL where STL, well-suited for interpretable behavioral descriptions, is used directly for event generation, while RM serve to define task goals in a modular and operationally effective manner. Furthermore, we leverage efficient methods for online monitoring of STL specifications implemented in the {\rlrom} framework \footnote{\url{https://github.com/decyphir/rlrom}} for training models that perform complex tasks or reliably converge toward behaviors satisfying specified requirements.
  
Our contributions can be summarized as follows:
\begin{itemize}
    \item We introduce a new framework, called \RMSTL{}, for RM incorporating STL predicates. The framework allows multiple RM to be composed in parallel, where the overall reward is computed as a weighted sum of the individual RM rewards. 
\item We propose an implementation of the framework, which has been successfully applied for a number of case studies. 
\end{itemize}

The paper is organized as follows. We first review the problem of RL with RM. Then we describe our contributions: an extension of RM with STL predicates, and a learning framework for such RM. We finally describe our implementation and demonstrate the approach on a minigrid, a cart-pole and highway-env examples with non-trivial tasks. 

\section{Preliminaries}

\subsection{Reinforcement learning problem}

The RL problem is formalized using an MDP in form of a tuple $M = (S,A,r,p,\gamma)$ where $S$ is a set of environment states, $A$ is a finite set of actions, $r : S \times A \times S \to \mathbb{R}$ is a reward function, $p(s_{t+1}|s_t,a_t)$ is a transition probability distribution, and $\gamma \in (0, 1]$ is a discount factor. Some states can be labeled as {\em terminal}.
The learning process is organized in episodes. One episode is defined as a sequence of interactions between an agent and its environment that starts from an initial state and ends when a terminal state is reached or a stopping condition is satisfied. 
At each time step $t$ of an episode, observing that the environment is at a state $s_t \in S$, the agent selects and executes an action according to a probability distribution over the actions $\pi(\cdot~|~s_t)$, called a policy. We assume in this work that the agent can observe the full state of the environment. The environment moves to a new state $s_{t+1}$ according to $p(\cdot~|~s_t, a_t)$ and the agent receives a reward $r(s_t, a_t, s_{t+1})$. 
The objective of the agent is to find a policy $\pi^*$ to maximize the expected discounted future reward from any state in $S$. Every optimal policy $\pi^*$ is known to satisfy the Bellman equations for every state $s \in S$ and every action $a \in A$: 
$q^{{\pi}^*}(s, a) = \sum_{s'\in S} p(s'~|~s, a) r(s,a,s') + \gamma \max_{a'\in A} q^{{\pi}^*}(s', a')$,
where  $q^{\pi} (s, a)$ is the $q$-function defined as the expected discounted future reward for taking the action $a$ from the state $s$ and then following the policy $\pi$. 

\subsection{Reward machines}

The reward function in the RL framework is a way to reflect the tasks that the agent is expected to perform. However, being defined only over the states and actions, reward functions can not capture complex non-Markovian tasks which are history dependent. This motivated the introduction of RM which are based on finite state machines~\cite{Icarte2020}.
As an agent interacts with the environment moving from state to state, it also transitions between states within an RM, which is determined by high-level events from the environment that the agent can detect. To specify such events, it is often assumed that a set of propositions $P$  and a labeling function $L : S \times A \times S \rightarrow 2^P$ are given. For an experience $(s, a, s')$ (where $s'$ is the resulting state after executing the action $a$ from the state $s$), the labeling function assigns truth values to the propositional symbols in $P$. These truth assignments are inputs to the RM. 

\begin{definition}[RM]\label{defRM} Given a set of propositional symbols $P$, a set of environment states $S$, and a set of actions $A$, a RM is a tuple $RM_{P,S, A} = \langle U, u_0, \delta_u, \delta_r\rangle$ where $U$ is a finite set of RM states, $u_0 \in U$ is an initial RM state, $\delta_u: U \times 2^P \rightarrow U$ is the RM state-transition function, and  $\delta_r: U \times U \rightarrow [S \times A \times S \rightarrow \mathbb{R}]$ is the transition-reward function.
\end{definition}

The RM $\RM_{P,S,A}$ starts in the initial RM state $u_0$. At every step $t$, the RM
receives as input a truth assignment $\sigma_t$ which contains all the propositions in $P$ that are true in $(s_t,a, s_{t+1})$. The rewards are Markovian with respect to $(s_t, u_t) \in S \times U$; however, the RM state $u_t$ at time step $t$ depends not only on the current environment state $s_t$ but also on the history of the environment evolution. Hence, using such a RM one can specify non-Markovian reward functions~\cite{Icarte2018}. Some RM states can also be labeled terminal. 

Furthermore, as we will see, in order to incorporate STL predicates as RM propositions, we need to keep the history of the environment evolution.  
To this end, we define an experience sequence of length $t>0$, denoted by $\traj_t = s_1, a_1, \ldots, a_{t-1}, s_t$ where $s_1$ is an initial state of the environment. Let $\Traj$ be the sets of experience sequences that the environment can perform. For a given set of propositions $P$, we extend the definition of labeling function to $L: \Traj \to 2^P$.



This following definition is for MDP coupled with multiple RM.

\begin{definition}[MDP-RM]\label{def:MDPRMs} Given an MDP $\mathsf{M} = (S,A,p,\gamma)$, a set of propositional symbols $P$, a labeling function $L$, and a set of RM $\{ \RM^1, \ldots, \RM^m\}$ where $\RM^i_{P,S,A} = \langle U, u_0, \delta^i_u, \delta^i_r\rangle$ with $i \in \{ 1, \ldots, m \}$, we define $\mathsf{M}_{RMs} = (S,A,p,\gamma,P,L,U,u_0,\{\sigma^1_u, \ldots, \sigma^m_u \}, \{\sigma^1_r, \ldots, \sigma^m_r \})$.
\end{definition}


The execution of the environment coupled with the RM and the associated reward are defined as follows. At a time step $t>0$, 

\begin{itemize}
    \item an action $a \in A$ is computed based on some policy depending on the aggregated state $\vs = (s, \vu)$ where $s$ is the current environment state, and $\vu = (u_1, \ldots, u_m)$ is the vector of the current RM states;
\item this action $a$ leads the environment to the next environment state $s'$, as defined by the MDP; 
\item the experience sequence is updated by concatenating it with the new action and state: $\traj = \traj \cdot (a,s')$;
\item based on the labeling function over the updated experience sequence, the next state $u'_i$ for each RM $\RM^i$ is determined: $u_i' = \delta^i_u (u_i, L(\traj))$; 
\item the overall reward associated with this step is then defined as:
$$\rewardStep_t = \sum_{i \in {1,\ldots,m}} w_i \delta_r^i(u_i, u'_i)$$
where $w_i \in \reals$ is a weight associated with the RM $\RM^i$.
\end{itemize}

The policy of selecting actions is part of the chosen training algorithm. Note that any standard RL method is supported by our framework. Indeed, the aggregate state of the environment and the RM  state is used to define the observation space. In this way, the value functions computed by the training algorithms can explicitly account for the reward mechanisms captured by the RM. We remark that in previous work on RL with RM~\cite{Icarte2018}, separate value functions are typically computed for each RM state. By contrast, using the aggregate state representation allows for a more compact and unified encoding of the value functions. However comparing generic vs specific training methods for RL with RM is beyond the scope of this paper.

\section{Reward machines with STL (\RMSTL{})}
\subsection{STL specifications} 
STL is a logic typically interpreted over dense-time signals that take values in a continuous metric space~\cite{MalerN04}. STL uses signal predicates $\mu$ over a signal $y$ of the form $f(y(t)) \sim 0$ where $f$ is a scalar-valued function and $\sim \in \{ <,\leq, >, \geq, =, \neq \}$. It is then defined recursively using Boolean combinations of subformulas, or by applying an interval-restricted temporal operator (such as ``always'' ($\G$), ``eventually'' ($\F$), and ``until'' ($\U$) which have the usual meaning) to a subformula. If $I$ is a time interval, the syntax for STL is defined as:
$\varphi ~ :=  \true \; | \; \mu \; | \; \neg \varphi \; | \;
\varphi_1 \wedge \varphi_2 \; | \; \varphi_1 \U_I \varphi_2$.
The ``eventually'' $\F$ operator is defined as $\F_I \varphi \triangleq \true \U_I \varphi$, and the ``always'' $\G$ operator is defined as $\G_I \varphi \triangleq \neg (\F _I \neg \varphi)$. 
The semantics are described informally as follows.
The signal $y$ satisfies $f(y)> 0$ at time $t$ if $f(y(t))>0$. It satisfies $\varphi = \G_{(0,1]}(f(y)=0)$ if for all time $0<
t \leq 1$, $f(y(t))=0$. The signal satisfies $\varphi= \F_{[1,2)}(f(y)<0)$ iff there exists a time $t$ such that $1\leq t < 2$ and $f(y(t))<0$. 

\paragraph{Quantitative semantics} We use the quantitative semantics of STL~\cite{DM10} defined as robustness of satisfaction, to provide a fine-grained exploration. Given a signal $y$ and an STL formula $\varphi$, a function $\rho$ is defined such that when $\rho(\varphi,y,t)$ is positive it indicates that $(y,t)$ satisfies
$\varphi$, and its absolute value estimates the \emph{robustness} of
this satisfaction. If $\varphi$ is of the form
$f(y)>b$, then its robustness is $\rho(\varphi,y,t) = f(y(t))-b$.  
For the conjunction $\varphi := \varphi_1 \wedge \varphi_2$, we have
$\rho(\varphi,y,t)=\min \left( \rho(\varphi_1,y,t),\rho(\varphi_2,y,t)\right)$,
while for the disjunction $\varphi := \varphi_1 \vee \varphi_2$, 
$\rho(\varphi,y,t)=\max\left(\rho(\varphi_1,y,t),\rho(\varphi_2,y,t)\right)$.
For a formula with until operator $\varphi := \varphi_1 \U_I \varphi_2$,
$\rho(\varphi,y,t) = \max_{t^\prime\in
  I}\left(\min\left(\rho(\varphi_2,y,t^\prime),\min_{t^{\prime\prime}\in
  [t,t^\prime]}\left(\rho(\varphi_1,y,t^{\prime\prime})\right)\right)\right)$.

STL satisfaction robustness is defined for signals over some given horizon. In order to use it in the RL framework where we need to deal with partial signal representing the evolution of the environment and the RM before the end of an episode, we use the notion of robustness interval, introduced in~\cite{Deshmukh2017} for online monitoring purposes, to enclose the exact robustness value that cannot yet be computed since the remaining portion of the signal is unknown.
In~\cite{Deshmukh2017} an algorithm is proposed to compute the robustness interval $[\minrho,\maxrho]$ for a given STL formula $\spec$ and a given partial signal. 

In a standard reinforcement learning setting, an agent interacts with its environment at discrete time steps, consequently, signals are observed in discrete time. We use an interpolation to obtain continuous-time signals. For simplicity, we use the same notation to mean both a signal in discrete time 
and its associated continuous-time signal.



\subsection{Using STL with RM}

We use the RM structure from Definition~\ref{def:MDPRMs}, while enriching the set $P$ of propositions with formulas involving STL robustness of the form $\minrho(\spec, \outsig_{[0,\tau]}, t) \geq \beta$ or $\maxrho(\spec, \outsig_{[0,\tau]}, t) \leq \beta$ where $t \leq \tau$ and $\spec$ is an STL formula. Here, $\outsig_{[0,\tau]}$ represents the evolution of the observation from the initial step to step $\tau$. It is possible that the formulas involve only a subset of variables in the observation. We refer to the STL formulas used to define RM transitions as {\em event formulas}. 


The resulting MDP-RM works as follows. At a time step $t$, let $\traj_t = s_0,a_1,s_1,\ldots,s_{t-1}, a_{t-1}$ be the sequence of experiences of the environment up to time step $(t-1)$.  Let $\vu_{t-1}$ be the vector of the current states of the RM. For each proposition $\varphi$ in $\Props$ that involves STL robustness, to determine its truth value we evaluate the interval of values for each robustness variable in the formula. Let $\sigma_t$ be a set of all the formulas in $\Props$ that are true at time step $t$. The truth assignment $\sigma_t$ is then received by the RM to decide the transitions to take from the current RM states $\vu_t$.

Besides event formulas, STL can be used to specify non-functional properties, which we refer to as {\em evaluation formulas}. These formulas are assessed over complete episodes to measure the agent’s performance. A straightforward approach might incorporate the online robustness of evaluation formulas into the observation space or use it to shape the reward signal as in \cite{hamilton2022sefm}. However, because evaluation formulas typically span long horizons—designed to evaluate complete episodes—their instantaneous robustness values offer limited insight into overall satisfaction, which can only be determined at the end of each episode. This poses a challenge for preserving the Markov property of the reward. For precisely this reason, it is also preferable to use short-horizon event formulas, where online robustness is more informative, and in this case the observation of the training algorithm can be additionally augmented with the robustness variables to encode in the value function the information about important events. These short-horizon formulas are assessed in a sliding-window fashion If $hz$ is the horizon of the formula, this means that we use as observation $\rho(\varphi, y, t^{-hz})$ where $t^{-hz}= \max(0, t-hz)$.

\subsubsection*{Discussion}
Before continuing, we briefly discuss the advantage of the proposed combination \RMSTL. The motivation for this extension, as mentioned earlier, is to directly capture events that depend on the evolution of the environment and RM states. These events are significant because they relate either to the reward mechanism or to the training guidance. Such guidance can steer the training algorithm toward exploring behaviors that are promising for optimizing the expected future reward, or prevent it from persisting in blocking behaviors that arise when the environment becomes trapped near a local optimum—situations that can be effectively characterized using STL. In these cases, the stopping condition can be encoded as a transition leading to a terminal state. These uses of STL with RM will be illustrated in the case studies.

Another advantage of this combination is interpretability. Using STL predicates to define transitions of RM and monitoring their robust satisfaction during the training process provide an expressive and yet interpretable means of modeling sophisticated reward mechanisms that can differentiate among subtle behavioral contexts. 

\section{EXPERIMENTAL RESULTS}
To evaluate {\RMSTL} on complex tasks within RL environments, we present three distinct environments as case studies: Minigrid, Cart-pole and Highway-env. In this section, we first describe the implementation, then each environment and the design of their specific tasks, and finally, the experimental results.

\subsection{Implementation}\label{sec:impl} 
The implementation of our framework \RMSTL{} builds upon the  \rlrom{} framework\footnote{\url{https://github.com/decyphir/rlrom}}, which can evaluate RL agents through interpretable monitors. \rlrom{} receives STL predicates as inputs, which are used to monitor the behavior of the agent and guide the training process. 
Within this framework, the environment is wrapped sequentially in two layers: 
\begin{enumerate}
\item an STL wrapper that evaluates the truth value and robustness of the user-defined STL predicates at each step of the episode. This wrapper serves as a real-time logical monitor, computing satisfaction degrees that reflect how well the behavior of the agent aligns with temporal and logical constraints ({\em e.g.}, ``the agent must pick up the key before unlocking the door'', or ``the car must keep a safe distance from the closest car''). The STL robustness evaluation over partial signals is based on STLRom\footnote{\url{https://    github.com/decyphir/stlrom}} which provides online robust monitoring as described in~\cite{Deshmukh2017}. 
 
\item an RM wrapper that constructs the reward signal according to the RM transitions by augmenting the observation space with the RM states and STL robustness, producing rewards consistent with the RM structure.
\end{enumerate} 
The two wrappers operate hierarchically, first applying logical monitoring via STL then the RM to structure task execution. This hierarchical design results in a flexible and interpretable framework: flexible, because STL predicates can be easily redefined or combined to express new temporal or logical constrains without modifying the underlying environment or learning algorithm; and interpretable, because the RM explicitly encodes the progress of the agent through discrete task stages, making it possible to trace why rewards are assigned and how decisions relate to high-level objectives. Together, these properties enable the same framework to generalize across multiple environments while maintaining a clear correspondence between agent behavior and task logic. 

Concerning the RL environments in the case studies, they were implemented using Python Gymnasium libraries~\cite{gymnasiumLibrary} that provide a standardized API for defining and interacting with environments. 
All agents were trained using the Proximal Policy Optimization (PPO) algorithm implemented in Stable-baselines3~\cite{stable-baselines3} (SB3), though RLRom makes it possible to pick any algorithm from SB3. The hyperparameters used in this implementation are in Table \ref{table:ppo_hyperparams_minigrid} in the Appendix.

\subsection{Minigrid}
The first case study uses Minigrid library, which provides a collection of 2D grid-world environments with goal-oriented tasks \cite{Minigrid23}. We used \textit{Unlockenv} as a base environment, but its grid size was increased to test scalability, as illustrated in Figure \ref{fig:minigrid}. The objective of the agent is to pick up the key and then open the door. The agent in these environments is a triangle-like agent with a discrete action space, which consists of seven actions (0–6), enabling the agent to move (left, right, forward), to interact with objects ({\em e.g.}, key, lava, ball), to perform pickup and drop actions, and to use the toggle action to activate objects, such as opening a door when positioned in front of it. The reward of $1 - 0.9 \frac{n}{n_{max}}$ is given for success, and $0$ for failure (where $n$ is the number of steps until success, and $n_{max}$ is the maximal number of steps of an episode defined as a function of the grid size). The episode ends if the agent opens the door or timeout, that is $n_{max}$ steps are reached. The observation space of the environment is partial and egocentric. It is represented as a $7\times7$ grid centered around the current position and orientation of the agent. Each cell in the grid is described by a 3-dimensional tuple: (\textit{object\_idx, color\_idx, state}), where \textit{object\_idx} and \textit{ color\_idx} correspond to the type and color of the objects observed by the agent, respectively. The \textit{state} component indicates the status of the object ({\em e.g.}, door open or closed). The objects included in this environment are the agent, a key, a door and surrounding walls that define the grid boundaries.

The objective of the agent can be decomposed into two sequential tasks: first, picking up the key, and then opening the door. Figure~\ref{fig:rm_minigrid} illustrates the corresponding RM, which structures the task into three states. The transitions between these states are governed by STL specifications, defined as $\varphi_1 ~ := \F_{[0,T]} has\_key$ and $\varphi_2 ~ := \F_{[0,T]} has\_key \ \land \ open\_door$. The value next to the specification corresponds to the reward associated with taking that transition. In this machine, $u_0$ denotes the initial state (before obtaining the key), $u_1$ represents the intermediate state where the agent holds the key, and $u_2$ corresponds to the terminal state, achieved when the agent successfully opens the door using the key.

\begin{figure}
    \centering
    \includegraphics[scale=0.3]{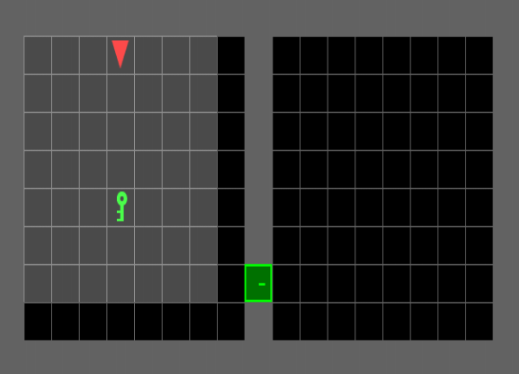}
    \caption{The minigrid environment \textit{Unlockenv} with bigger grid size.}
    \label{fig:minigrid}
\end{figure}
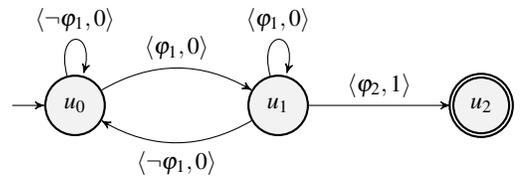
\begin{figure}[htbp]
    \centering
    \begin{tikzpicture}[transform shape, scale=.9]
        \tikzset{
        ->,
        >=stealth',
        every state/.style={thick, fill=gray!10},
        initial text=
        }
        \node[state, initial](u0){$u_0$};
        \node[state] at (3,0)(u1){$u_1$};
        \node[state, accepting] at (6,0)(u2){$u_2$};

        \draw (u0) edge[bend left, above] node{$\langle \varphi_1, 0 \rangle$} (u1);
        \draw (u1) edge[bend left, below] node{$\langle \neg \varphi_1 , 0 \rangle$} (u0);
        \draw (u0) edge[loop above] node{$\langle \neg \varphi_1 , 0 \rangle$} (u0);
        \draw (u1) edge[loop above] node{$\langle \varphi_1 , 0 \rangle$} (u1);
        \draw (u1) edge[right, above] node{$\langle \varphi_2, 1\rangle$} (u2);
       
    \end{tikzpicture}
    \caption{Reward machine of the Minigrid agent}
    \label{fig:rm_minigrid}
\end{figure}

\begin{figure}[t]
    \centering
    \begin{subcaptionbox}{Average episode reward.\label{fig:a}}
        {\includegraphics[scale=0.25]{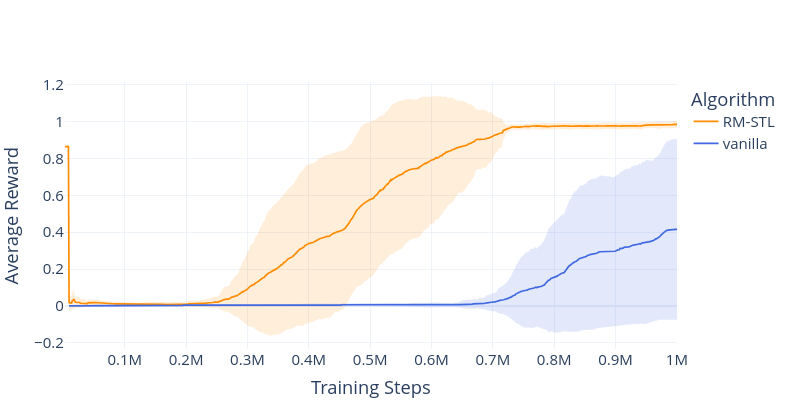}}
    \end{subcaptionbox}
    \hfill
    \begin{subcaptionbox}{Average episode length.\label{fig:b}}
        {\includegraphics[scale=0.25]{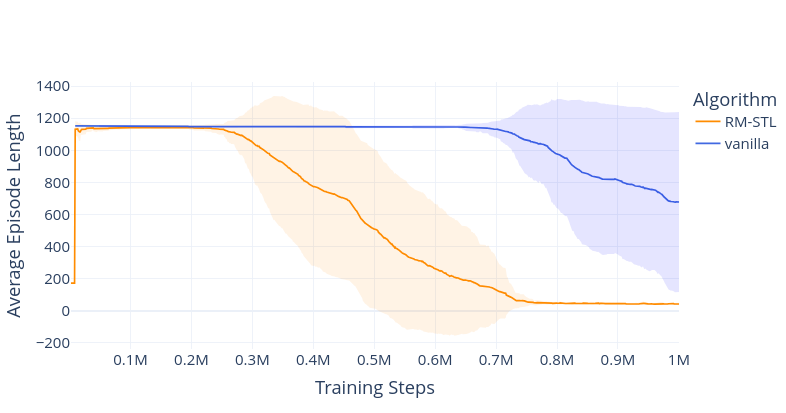}}
    \end{subcaptionbox}

    \caption{Results from training an agent in the \textit{Unlock} environment from a extended grid size 12x12 in MiniGrid. Blue: vanilla; orange: RM state in observation with STL specification during training.}
    \label{fig:minigrid_training}
\end{figure}

The training was conducted on environments of increasing complexity, starting with grid size 6x6 and progressively expanding up to $12 \times 12$. For each grid configuration, two models were trained: a baseline vanilla PPO agent and the proposed \RMSTL{} implementation, under identical hyper-parameters 5 times each. Figure~\ref{fig:minigrid_training} corresponds to the results of the largest grid, illustrating the most challenging scenario.
These results include two key performance metrics, episode reward vs number of training time steps, and episode length vs number of training time steps, allowing a comprehensive comparison of both approaches across different environment complexities. 

\begin{figure}[t]
    \centering
    
    \begin{subcaptionbox}{Average episode reward.\label{fig:a.}}
        {\includegraphics[scale=0.3]{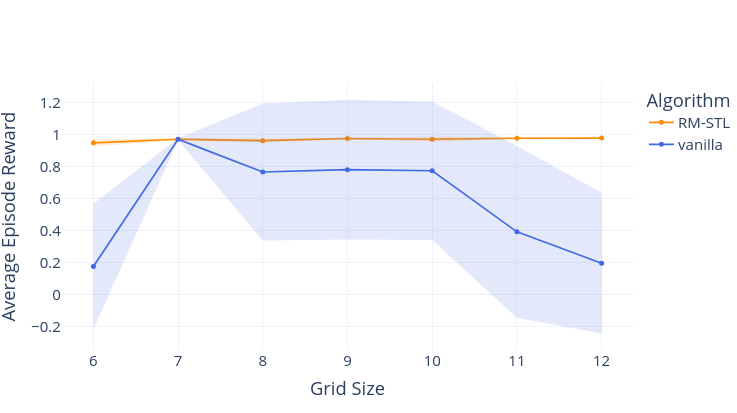}}
    \end{subcaptionbox}
    \hfill
    \begin{subcaptionbox}{Average episode length.\label{fig:b.}}
        {\includegraphics[scale=0.3]{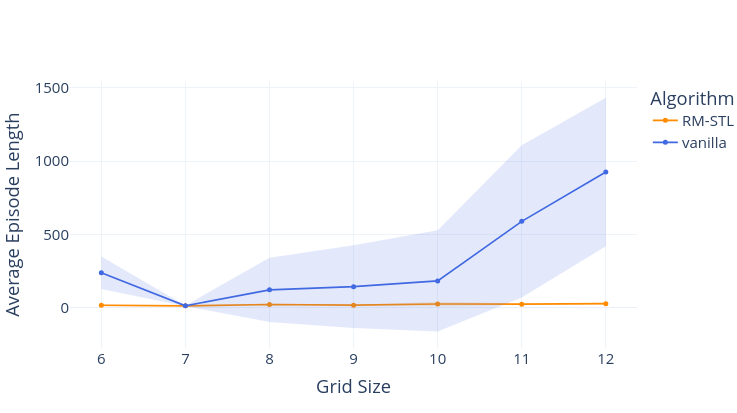}}
    \end{subcaptionbox}

    \caption{Results achieved by the trained models when evaluated on larger grid sizes.}
    \label{fig:minigrid_grid_comparison}
\end{figure}

The plot in Figure~\ref{fig:minigrid_grid_comparison}, where the x-axis represents the grid size and the y-axis the corresponding averaged metrics, demonstrates that the \RMSTL{} agent consistently outperforms the vanilla PPO as the environment complexity increases, achieving higher rewards and shorter episode lengths. More details of the results are shown in Table \ref{table:minigrid_results}. This trend highlights the improved learning efficiency and generalization capabilities provided by integrating STL specifications and RM into the training process.

\begin{table}[t]
\caption{Performance comparison per grid size of RM+STL vs Vanilla}
\label{table:minigrid_results}
\centering
\small
\resizebox{\columnwidth}{!}{
\begin{tabular}{|c||c|c|c|}
\hline
Grid Size & Algorithm & Mean Reward $\pm$ Std & Mean Len Episode $\pm$ Std \\
\hline
6  & Specs   & $0.9494 \pm 0.0221$ & $16.2 \pm 7.09$ \\ \hline
6  & Vanilla & $0.1756 \pm 0.3927$ & $238.2 \pm 111.36$ \\ \hline
7  & Specs   & $0.9720 \pm 0.0062$ & $12.2 \pm 2.68$ \\ \hline
7  & Vanilla & $0.9715 \pm 0.0060$ & $12.4 \pm 2.61$ \\ \hline
8  & Specs   & $0.9631 \pm 0.0128$ & $21.0 \pm 7.28$ \\ \hline
8  & Vanilla & $0.7670 \pm 0.4290$ & $121.2 \pm 218.63$ \\ \hline
9  & Specs   & $0.9767 \pm 0.0015$ & $16.8 \pm 1.10$ \\ \hline
9  & Vanilla & $0.7817 \pm 0.4370$ & $142.8 \pm 282.42$ \\ \hline
10 & Specs   & $0.9721 \pm 0.0171$ & $24.8 \pm 15.24$ \\ \hline
10 & Vanilla & $0.7748 \pm 0.4335$ & $182.4 \pm 345.66$ \\ \hline
11 & Specs   & $0.9782 \pm 0.0050$ & $23.4 \pm 5.41$ \\ \hline
11 & Vanilla & $0.3926 \pm 0.5375$ & $588.8 \pm 519.24$ \\ \hline
12 & Specs   & $0.9789 \pm 0.0055$ & $27.0 \pm 7.00$ \\ \hline
12 & Vanilla & $0.1969 \pm 0.4402$ & $925.6 \pm 506.25$ \\ \hline
\end{tabular}}
\end{table}

\subsection{Cart-Pole}
\def\stuck{\varphi_{\text{stuck}}}
The second case study involves a cart-pole environment, where a pole is attached by an unactuated joint to a cart which moves along a friction-less track. The pole is placed upright on the cart and the goal is to balance the pole by applying forces in the left and right direction on the cart. The default reward is $1$ for every step taken if the pole is upright. The state space is continuous and has $4$ dimensions (cart position, cart speed, pole angle, pole angular speed). The agent has $2$ available discrete actions: push the cart to the right or to the left. The episode ends if the absolute value of pole angle is greater than $12^{o}$, or the absolute value of the cart position is greater than $2.4$, or if the episode length is greater than $500$ time steps~\cite{Farama2023}.

\begin{figure}
    \centering
    \includegraphics[scale=0.25]{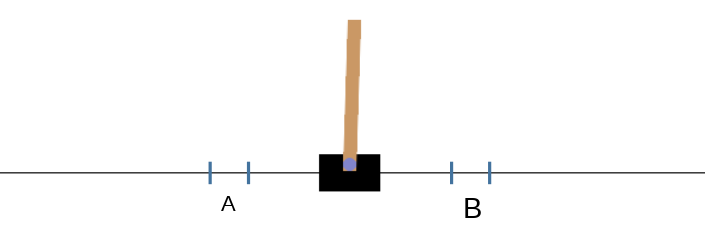}
    \caption{The cart-pole environment, augmented with target regions
      A and B defined as STL predicates $\mu_A$ and $\mu_B$.}
    \label{fig:cartpole_ab}
\end{figure}

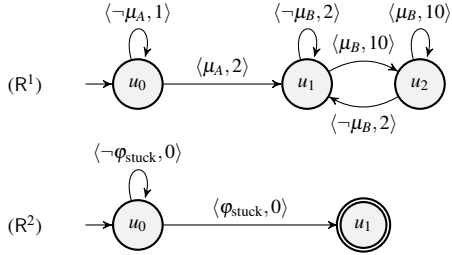
\begin{figure}[htbp]

  \centering
  \begin{tikzpicture}[transform shape, scale=.75]
        \tikzset{
        ->,
        >=stealth',
        every state/.style={thick, fill=gray!10},
        initial text=
      }

        \node (R) at (0,0) {($\RM^1$)};
        \node[state, initial] at (2,0)(u0){$u_0$};
        \node[state] at (5,0)(u1){$u_1$};
        \node[state] at (7,0)(u2){$u_2$}; 
        \draw (u0) edge[loop above] node{$\langle \neg \mu_A, 1\rangle$} (u0);
        \draw (u0) edge[left, above] node{$\langle \mu_A, 2 \rangle$} (u1);
        \draw (u1) edge[loop above] node{$\langle \neg \mu_B, 2\rangle$} (u1);
        \draw (u1) edge[above, bend left] node{$\langle \mu_B , 10 \rangle$} (u2);
        \draw (u2) edge[loop above] node{$\langle  \mu_B, 10 \rangle$} (u2);
        \draw (u2) edge[below, bend left] node{$\langle \neg \mu_B, 2 \rangle$} (u1);

        \node (R) at (0,-2.5) {($\RM^2$)};
      
        \node[state, initial] at (2,-2.5)  (u0){$u_0$};
        \node[state, accepting] at (6,-2.5)(u1){$u_1$};
 
        \draw (u0) edge[loop above] node{$\langle \neg \stuck, 0\rangle$} (u0);
        \draw (u0) edge[left, above] node{$\langle \stuck, 0 \rangle$} (u1);

      \end{tikzpicture}

      \caption{Reward machines of the cart pole: $\RM^1$
         encodes the task to go to
      region A and then to region B and stay there. $\RM^2$ enforces episode termination when the formula $\stuck$ defined in formula~\eqref{cartTermi} is true.}
    \label{fig:rm_cartpole}
\end{figure}



To make more complex tasks in this environment, specific target positions are defined. Figure~\ref{fig:cartpole_ab} illustrates the initial position of the cart-pole system, along with the designated target regions. The first objective of the cart is to reach the region $A$, where the 
$x$ position must be between $-0.7$ and $-0.5$ (formally: $\mu_A= -0.7 < x <  -0.5 $). Subsequently, the cart has to move to the region $B$, located between $0.5$ and $0.7$ (formally: $\mu_B = 0.5 < x <  0.7 $), all while keeping the pole upright throughout the episode. This task is significantly more difficult than the original cart-pole task because in addition to not falling, the agent has to perform a sequence of subtasks (move left, move right, and then brake). It is not easy to use simple reward functions to specifying such a task composed of three subsequential subtasks, since the reward should reflect which subtasks have been achieved and which remain to be executed. The RM describing this objective is presented in Figure~\ref{fig:rm_cartpole}, where the goal is to reach the state $u_2$ and stay there for the length of the episode. To keep the incentive to not fall, a minimum reward of $1$ is provided for each step, as in the original reward formulation. This reward is doubled when the cart-pole has reached the region A, and then the reward becomes $10$ for each step whenever the cart-pole is in B.   

\begin{figure}[htbp]
  \centering
    \includegraphics[scale=0.35]{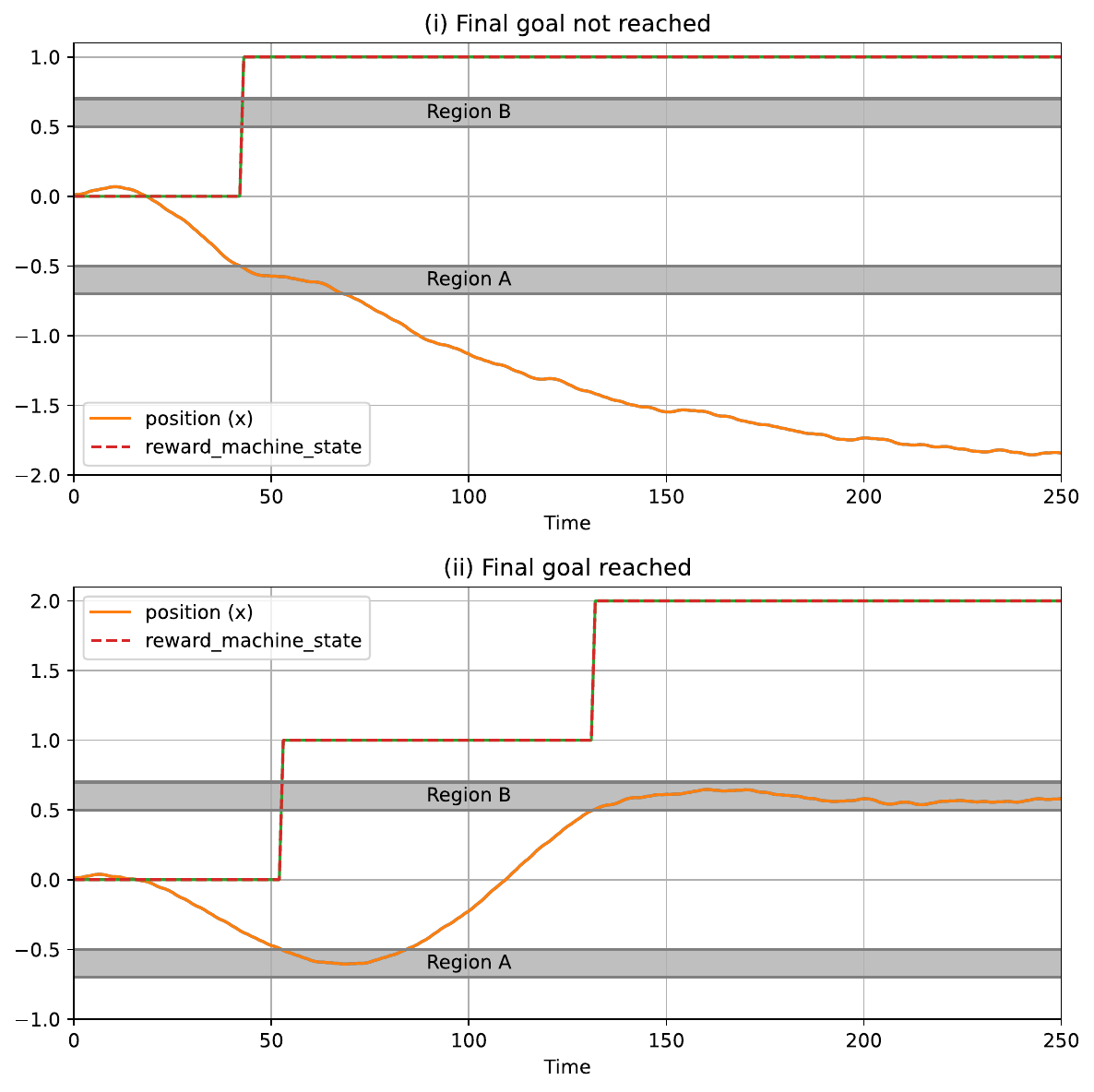}
    \caption{Test results of a single episode using the sub-optimal trained model (i) and using a successfully trained one (ii).}
    \label{fig:good_vs_bad}
\end{figure}

\begin{figure}[htbp]
    \centering
    \includegraphics[scale=0.45]{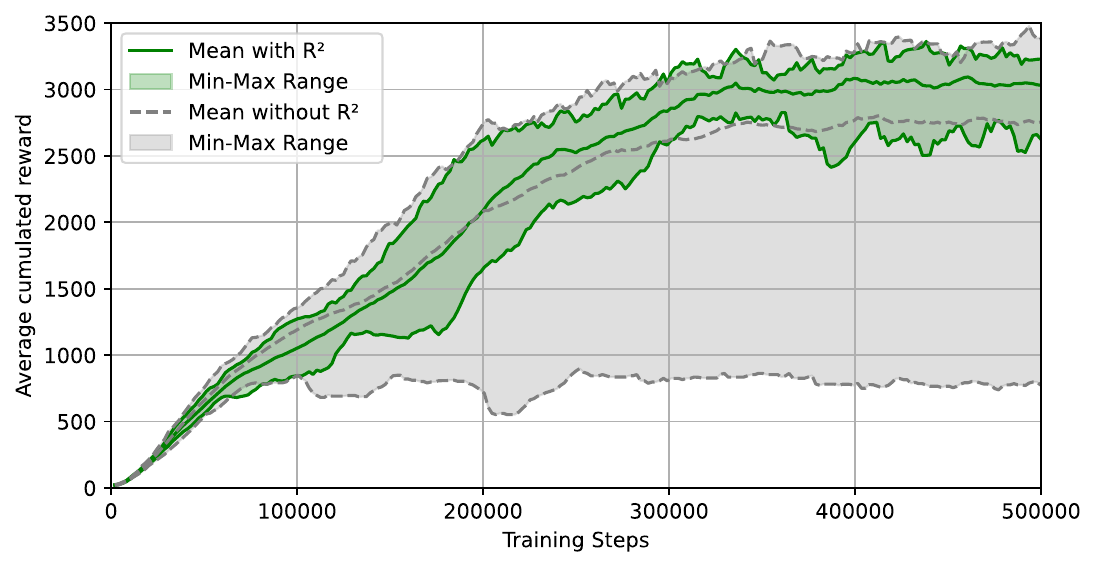}
    \caption{Results for 10 training instances with (green) and without (gray) $\RM^2$ encoding STL formula $\stuck$ as terminal condition. With $\RM^2$, the performance of the policy consistently reaches an accumulated reward of 3.000, indicating that the task is successfully completed,  whereas with only $\RM^1$, the performance of some policies remains around 800, indicating that only the region A is reached in most episodes.}
    \label{fig:training_with_STL}
\end{figure}

Figure~\ref{fig:rm_cartpole} also depicts another RM for the cart-pole which encodes a training guidance. Indeed, after reaching the region A, the agent may not move to B because it can still get a reward of $2$ at each step. In order to encourage the exploration towards B, we describe, by the STL formula~\eqref{cartTermi}, situations where at some time point (in this example, between the $100^{th}$ step and the $200^{th}$ step) the cart continuously stays in the region with $x<0$ for some time (in this example, for $300$ successive steps).   
\begin{equation}
\label{cartTermi}
\stuck = \F_{[100, 200]} \G_{[0,300]} (x<0)
\end{equation}
Using this in a RM with a terminal state penalizes the sub-optimal policy illustrated on Figure~\ref{fig:good_vs_bad} (i). Figure~\ref{fig:training_with_STL} shows the results of $10$ instances training first without $\RM^2$, then with $\RM^2$. We show the mean average cumulated reward, as well as the average cumulated reward of the worst and the best policies obtained both without $\RM^2$ and with $\RM^2$. For the latter case, {\em i.e.}, using $\RM^2$, we see that the sub-optimal policy is completely eliminated, as the average cumulated reward eventually reaches values around 3.000 for all training instances. 

\subsection{Highway-env}

\def\f{\varphi}
\def\fa{\mu_{fast}}
\def\ri{\mu_{right}}
\def\nfa{\neg \mu_{fast}}
\def\saf{safedist}
\def\nsaf{\neg safedist}

\def\statesafe{$u_{safe}$}
\def\stateunsafe{$u_{danger}$}
\def\lazy{\f_{lazy}}
\def\mulazy{\mu_{lazy}}
\def\tail{\f_{tail}}
\def\evalfast{\varphi_{fast}}
\def\phileft{\varphi_{left}}
\def\evalright{\varphi_{right}}
\def\danger{\mu_{danger}}
\def\bv{\rule{0pt}{2ex}}

\begin{figure}
  \includegraphics[width=.5\textwidth]{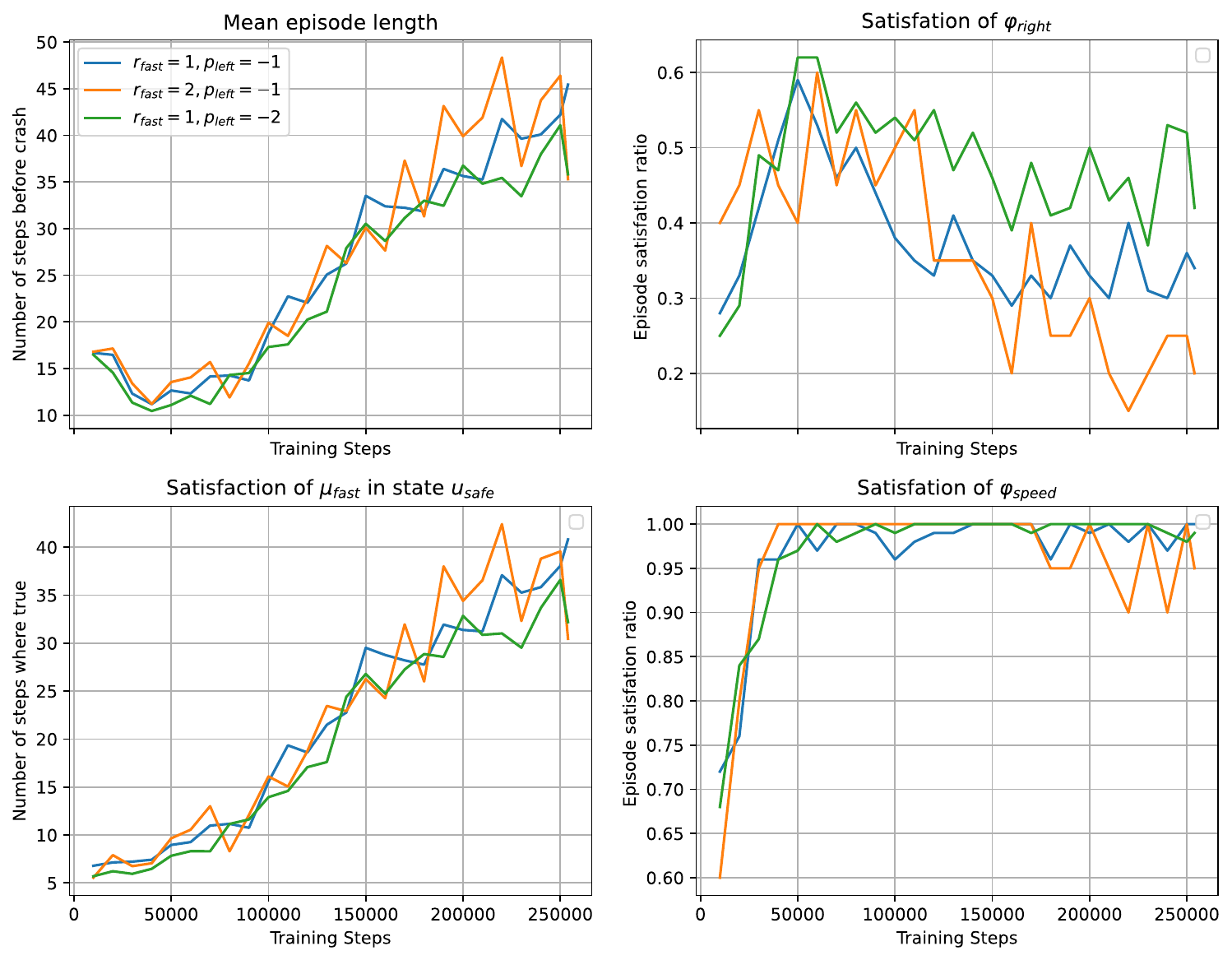}
  \caption{Training results for different values of rewards and penalties. Every 1e5 steps, the agent is tested for 100 episodes and all formulas are evaluated to give a complete picture of the current performance. All values lead to improving performance, lower values of $p_{left}$ increase the rate of satisfaction of $\evalright$, which is consistent with the formulas and RM.}
  \label{fig:hwtrainings}
\end{figure}

Highway-env is a minimalist environment for decision-making in autonomous driving. It includes basic driving scenarios such as highways, intersections, roundabouts, racetracks, and parking setups, where the agent can perform different tactical decision making tasks in the RL context. The vehicles populating the road follow simple and realistic behaviors that dictate how they accelerate and steer on the road. For this experiment, we use an environment where the observation has the position $x^{ego}$ and $y^{ego}$, the speed $v^{ego}_x$ and $v^{ego}_y$ of the ego vehicle, as well as position and speed for the four closest vehicles in front of it, denoted as $x^i$, $y^i$, $v^{i}_x$ and $v^{i}_y$ for $i\in\{1,2,3,4\}$. The available actions of the agent are: turn left or right, idle, go faster or slower. The original reward function consists of a velocity term and a collision term. In our experiment, we aim at training agents to not only drive fast and safe, but also to follow certain driving behaviors. For this, we designed a set of formulas, described in Table~\ref{table:hw-env} and RM described in Figure~\ref{fig:rm_hw-env}. They are designed to enforce three main goals: 1) avoid collision, 2) go fast and 3) prefer the right lane. These goals are often in competition with one another, but with our framework, we can balance them in a comprehensive way. For instance, high velocity should not be rewarded when the ego vehicle is close to a collision with another car, {\em etc}. By tuning the rewards and penalties associated with each transition, we can also influence which policy is prioritized wrt to one another. For illustration, in Fig~\ref{fig:hwtrainings}, we show the evolution of formula satisfaction and metrics during training (testing and monitoring 100 episodes every 1e5 steps of training), which is consistent with the definitions in Table~\ref{table:hw-env} and Fig~\ref{fig:rm_hw-env}.

\begin{table}
 
  \centering
 
  \begin{tabular}{|l|c|l|}

    \hline
 \bv  \textbf{Formula}                         & $hz$ & \textbf{Interpretation}  \\ \hline
 \bv    $\danger^i= |x^i|<.1 \wedge |y^i|<.1$   & 0     &car $i$ is in front and too close\\ 
 \bv    $\danger= \vee_{i=1}^4\ \danger^i$        & 0     &A car is in front and too close \\ 
 \bv    $\tail= \G_{[0,10]}\ \danger$             & 10    &Too close for too long  \\\hline
 \bv    $\fa =v^{ego}_x>25$                        & 0     &ego vehicle goes fast \\
 \bv    $\lazy= \G_{[0,10]}\ \neg \fa $             &  10    &safe for 10 steps \\ \hline
 \bv    $\evalfast$ = $\F_{[0, 85]} \G_{[0, 15]}\ \fa$&100  & always accelerates eventually\\ \hline  
 \bv    $\ri= y^{ego} > 0.6 $                       & 0     &ego is in right lane \\ 
 \bv    $\phileft = \G_{[0,10]}\ \neg \ri $          & 10   &not in right lane for 10 steps \\ 
 \bv    $\evalright$ = $\F_{[0, 85]} \G_{[0, 15]}\ \ri$&100 & always in right lane eventually\\ \hline  
 \end{tabular}

 \caption{Formulas characterizing driving behaviors and used in the RM in Figure~\ref{fig:rm_hw-env}. Short horizon formula ($hz\leq10$) are used as event formulas. Long horizon formulas ($hz=100$) are for agent evaluation past training.}
 \label{table:hw-env}
\end{table}

\def\rewA{\textcolor{green}{r_{fast}}}
\def\rewB{\textcolor{green}{r_{1}}}
\def\penLazy{\textcolor{red}{-5}}
\def\penfada{\textcolor{red}{-1}}
\def\pentail{\textcolor{red}{-1}}
\def\penleft{\textcolor{red}{p_{left}}}
\def\pennfanda{\textcolor{red}{p_{left}}}

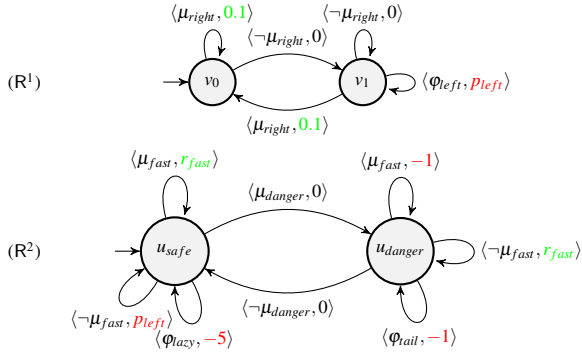
\begin{figure}[htbp]
    \centering

\begin{tikzpicture}
        \tikzset{
        ->,
        >=stealth',
        every state/.style={thick, fill=gray!10},
        initial text=
      }[transform shape, scale=.75]
\scriptsize
\node (R2) at (-2,0){($\RM^2$)};
\node[state, initial, minimum width=.9cm](u0){\statesafe};
\node[state] at (3,0)(u1){\stateunsafe};
\draw (u0) edge[loop above] node                        {$\langle \fa,          \rewA \rangle$} (u0); 
\draw (u0) edge[out=300,in=270,looseness=8] node[below] {$\langle \lazy, \penLazy     \rangle$} (u0); 
\draw (u0) edge[out=240,in=210,looseness=8] node[below] {$\langle \neg \fa,     \pennfanda \rangle$} (u0); 
\draw (u0) edge[bend left, above] node                  {$\langle \danger,      0     \rangle$} (u1); 
\draw (u1) edge[loop above] node                        {$\langle \fa,          \penfada \rangle$} (u1); 
\draw (u1) edge[out=300,in=270,looseness=8] node[below] {$\langle \tail,        \pentail \rangle$} (u1); 
\draw (u1) edge[loop right] node                        {$\langle \neg \fa,     \rewA \rangle$} (u1); 
\draw (u1) edge[below, bend left] node                  {$\langle \neg \danger, 0 \rangle$} (u0);                

\node (R1) at (-2,2.25){($\RM^1$)};
\node (v0) [initial, state] at (0.5, 2.25){$v_0$};
\node (v1) [state] at (2.5, 2.25){$v_1$};
\draw (v0) edge[loop above] node                        {$\langle \ri, \textcolor{green}{0.1} \rangle$}   (v0);
\draw (v0) edge[bend left, above] node                  {$\langle \neg \ri, 0 \rangle$}   (v1);
\draw (v1) edge[bend left, below] node                  {$\langle \ri, \textcolor{green}{0.1} \rangle$}   (v0);
\draw (v1) edge[loop above] node                        {$\langle \neg \ri, 0 \rangle$}   (v1);
\draw (v1) edge[loop right] node                        {$\langle \phileft, \penleft \rangle$}   (v1);

\end{tikzpicture}

\caption{Reward machine and formulas for the highway-env environment.}
    \label{fig:rm_hw-env}
\end{figure}


\section{Conclusion} 
In this paper, we proposed an extension of RM using STL predicates to incorporate strategies to guide RL. This  framework accelerates the training process by avoiding experiences that maintain the agent around local optima or are not robust with respect to safety specifications. We implemented our algorithms and demonstrated their effectiveness across three case studies.

Future work includes using STL for reward shaping. Similar to the essential ideas of the potential-based reward shaping approach \cite{Icarte2022}, but instead of treating RM as MDP, STL can be used to discover promising or, on the contrary, critical experiences to reflect them in intermediate rewards. This however should be done using a search procedure with good coverage over the behavior space. Another natural direction is STL shielding for safe-RL. Our combination of RM and STL provides a flexible and practical framework to restrict certain critical actions under specific conditions. For instance, in the highway-env case studies, agents should be strictly prohibited from accelerating when $\danger$ is true. Finally, we aim to extend our framework to recent variable-time-step RL approaches, where actions are executed only when necessary. In this context, the dense-time semantics of STL offers a distinct advantage over other discrete time temporal logics, as they naturally accommodate asynchronous decision-making and continuous-time reasoning.


\section{Acknowledgments}
This work is partially supported by the joint French-Japanese ANR-JST project CyPhAI, the Auvergne-Rh{\^o}ne-Alpes Region Project DetAI,  and by a French state grant managed by the National Research Agency as part of France 2030, with the reference ANR-23-IACL-0006.


\addtolength{\textheight}{-12cm}   

\bibliographystyle{ieeeconf}
\bibliography{ECC2026}

@INPROCEEDINGS{BeltaCDC2016,
  author={Aksaray, Derya and Jones, Austin and Kong, Zhaodan and Schwager, Mac and Belta, Calin},
  booktitle={2016 IEEE 55th Conference on Decision and Control (CDC)}, 
  title="{Q-Learning for robust satisfaction of Signal Temporal Logic specifications}", 
  year={2016},
  volume={},
  number={},
  pages={6565-6570},
  doi={10.1109/CDC.2016.7799279}
}

@article{Belta2017,
  author       = {Xiao Li and
                  Cristian Ioan Vasile and
                  Calin Belta},
  title        = {Reinforcement Learning With Temporal Logic Rewards},
  journal      = {CoRR},
  volume       = {abs/1612.03471},
  year         = {2016},
  url          = {http://arxiv.org/abs/1612.03471},
  eprinttype    = {arXiv},
  eprint       = {1612.03471},
  bibsource    = {dblp computer science bibliography, https://dblp.org}
}

@article{Brafman2018, 
title="{LTLf/LDLf Non-Markovian Rewards}",
volume={32},  
number={1}, 
journal={Proceedings of the AAAI Conference on Artificial Intelligence}, author={Brafman, Ronen and De Giacomo, Giuseppe and Patrizi, Fabio}, year={2018},
month={Apr.} 
}

@misc{Hasanbeig2019,
title={Reinforcement Learning for Temporal Logic Control Synthesis with Probabilistic Satisfaction Guarantees}, 
author={Mohammadhosein Hasanbeig and Yiannis Kantaros and Alessandro Abate and Daniel Kroening and George J. Pappas and Insup Lee},
year={2019},
eprint={1909.05304},
archivePrefix={arXiv},
primaryClass={cs.LO},
url={https://arxiv.org/abs/1909.05304}, 
}

@inproceedings{DeGiacomo2019,
author = {De Giacomo, Giuseppe and Iocchi, Luca and Favorito, Marco and Patrizi, Fabio},
year = {2021},
month = {05},
pages = {128-136},
title = "{Foundations for Restraining Bolts: Reinforcement Learning with LTLf/LDLf Restraining Specifications}",
volume = {29},
booktitle = {Proceedings of the International Conference on Automated Planning and Scheduling}
}

@INPROCEEDINGS{hamilton2022sefm,
  author={Hamilton, Nathaniel and Preston K. Robinette and Taylor T. Johnson},
  booktitle={20th International Conference on Software Engineering and Formal Methods (SEFM)}, 
  title={Training Agents to Satisfy Timed and Untimed Signal Temporal Logic Specifications with Reinforcement Learning}, 
  year={2022},
  month=sep,
  volume={},
  number={},
  pages={},
  doi={}
}

@InProceedings{Hahn2019,
author="Hahn, Ernst Moritz
and Perez, Mateo
and Schewe, Sven
and Somenzi, Fabio
and Trivedi, Ashutosh
and Wojtczak, Dominik",
editor="Vojnar, Tom{\'a}{\v{s}}
and Zhang, Lijun",
title="{Omega-Regular Objectives in Model-Free Reinforcement Learning}",
booktitle="Tools and Algorithms for the Construction and Analysis of Systems",
year="2019",
publisher="Springer International Publishing",
address="Cham",
pages="395--412"
}

@inproceedings{Hasanbeig2020,
author = {Mohammadhosein Hasanbeig and Daniel Kroening and Alessandro Abate},
year = {2020},
pages = {1-22},
title = "{Deep Reinforcement Learning with Temporal Logics}",
volume = {29},
booktitle = {Proceedings of 18th International Conference Formal Modeling and Analysis of Timed Systems FORMATS}
}

@article{Topcu2019,
  author       = {Zhe Xu and
                  Ufuk Topcu},
  title        = {Transfer of Temporal Logic Formulas in Reinforcement Learning},
  journal      = {CoRR},
  volume       = {abs/1909.04256},
  year         = {2019},
  url          = {http://arxiv.org/abs/1909.04256},
  eprinttype    = {arXiv},
  eprint       = {1909.04256},
  bibsource    = {dblp computer science bibliography, https://dblp.org}
}

@article{Kapoor2020,
  author       = {Parv Kapoor and
                  Anand Balakrishnan and
                  Jyotirmoy V. Deshmukh},
  title        = "{Model-based Reinforcement Learning from Signal Temporal Logic Specifications}",
  journal      = {CoRR},
  volume       = {abs/2011.04950},
  year         = {2020}
}

@INPROCEEDINGS{Kuo2020,
  author={Kuo, Yen-Ling and Katz, Boris and Barbu, Andrei},
  booktitle={2020 IEEE/RSJ International Conference on Intelligent Robots and Systems (IROS)}, 
  title="{Encoding formulas as deep networks: Reinforcement learning for zero-shot execution of LTL formulas}", 
  year={2020},
  volume={},
  number={},
  pages={5604-5610}
 }

@article{Topcu2021,
  author       = {Yuqian Jiang and
                  Sudarshanan Bharadwaj and
                  Bo Wu and
                  Rishi Shah and
                  Ufuk Topcu and
                  Peter Stone},
  title        = {Temporal-Logic-Based Reward Shaping for Continuing Learning Tasks},
  journal      = {CoRR},
  volume       = {abs/2007.01498},
  year         = {2020},
  url          = {https://arxiv.org/abs/2007.01498},
  eprinttype    = {arXiv},
  eprint       = {2007.01498},
  bibsource    = {dblp computer science bibliography, https://dblp.org}
}

@inproceedings{Vaezipoor2021,
  title="{LTL2Action: Generalizing LTL Instructions for Multi-Task RL}",
  author={Pashootan Vaezipoor and Andrew C. Li and Rodrigo Toro Icarte and Sheila A. McIlraith},
  booktitle={International Conference on Machine Learning},
  year={2021}
}

@InProceedings{Jothimurugan2022,
author="Kishor Jothimurugan and Rajeev Alur and Osbert Bastani",
title="{A composable specification language for reinforcement learning tasks}",
booktitle="Proceedings of the 33rd International Conference on Neural Information Processing Systems",
year="2019",
publisher="Springer Nature Switzerland",
address="Cham",
pages="13041--13051"
}

@article{Alur2022,
author       = {Rajeev Alur and
              Suguman Bansal and
              Osbert Bastani and
              Kishor Jothimurugan},
title        = {A Framework for Transforming Specifications in Reinforcement Learning},
journal      = {CoRR},
volume       = {abs/2111.00272},
year         = {2021},
url          = {https://arxiv.org/abs/2111.00272},
eprinttype    = {arXiv},
eprint       = {2111.00272},
bibsource    = {dblp computer science bibliography, https://dblp.org}
}

@InProceedings{Icarte2018,
title = 	 {Using Reward Machines for High-Level Task Specification and Decomposition in Reinforcement Learning},
author =       {Icarte, Rodrigo Toro and Klassen, Toryn and Valenzano, Richard and McIlraith, Sheila},
booktitle = 	 {Proceedings of the 35th International Conference on Machine Learning},
pages = 	 {2107--2116},
year = 	 {2018},
editor = 	 {Dy, Jennifer and Krause, Andreas},
volume = 	 {80},
series = 	 {Proceedings of Machine Learning Research},
month = 	 {10--15 Jul},
publisher =    {PMLR},
url = 	 {https://proceedings.mlr.press/v80/icarte18a.html}
}

@inproceedings{Icarte2019,
  title     = "{LTL and Beyond: Formal Languages for Reward Function Specification in Reinforcement Learning}",
  author    = {Camacho, Alberto and Toro Icarte, Rodrigo and Klassen, Toryn Q. and Valenzano, Richard and McIlraith, Sheila A.},
  booktitle = {Proceedings of the Twenty-Eighth International Joint Conference on
               Artificial Intelligence, {IJCAI-19}},
  publisher = {International Joint Conferences on Artificial Intelligence Organization},
  pages     = {6065--6073},
  year      = {2019}
}

@article{Icarte2020,
  author       = {Rodrigo Toro Icarte and
                  Toryn Q. Klassen and
                  Richard Anthony Valenzano and
                  Sheila A. McIlraith},
  title        = "{Reward Machines: Exploiting Reward Function Structure in Reinforcement
                  Learning}",
  journal      = {CoRR},
  volume       = {abs/2010.03950},
  year         = {2020}
}

@article{Icarte2022,
  author       = {Rodrigo Toro Icarte and
                  Toryn Q. Klassen and
                  Richard Anthony Valenzano and
                  Sheila A. McIlraith},
  title        = {Reward Machines: Exploiting Reward Function Structure in Reinforcement
                  Learning},
  journal      = {CoRR},
  volume       = {abs/2010.03950},
  year         = {2020},
  url          = {https://arxiv.org/abs/2010.03950},
  eprinttype    = {arXiv},
  eprint       = {2010.03950},
  bibsource    = {dblp computer science bibliography, https://dblp.org}
}

@article{Topcu2021rm,
author       = {Cyrus Neary and
              Zhe Xu and
              Bo Wu and
              Ufuk Topcu},
title        = {Reward Machines for Cooperative Multi-Agent Reinforcement Learning},
journal      = {CoRR},
volume       = {abs/2007.01962},
year         = {2020},
url          = {https://arxiv.org/abs/2007.01962},
eprinttype    = {arXiv},
eprint       = {2007.01962},
bibsource    = {dblp computer science bibliography, https://dblp.org}
}

@INPROCEEDINGS{MunirajCDC2018,
  author={Muniraj, Devaprakash and Vamvoudakis, Kyriakos G. and Farhood, Mazen},
  booktitle={2018 IEEE Conference on Decision and Control (CDC)}, 
  title="{Enforcing Signal Temporal Logic Specifications in Multi-Agent Adversarial Environments: A Deep Q-Learning Approach}", 
  year={2018},
  volume={},
  number={},
  pages={4141-4146},
  doi={10.1109/CDC.2018.8618746}
}

@INPROCEEDINGS{Ikemoto2022,
  author={Ikemoto, Junya and Ushio, Toshimitsu},
  booktitle={2022 IEEE 27th International Conference on Emerging Technologies and Factory Automation (ETFA)}, 
  title="{Deep Reinforcement Learning Based Networked Control with Network Delays for Signal Temporal Logic Specifications}", 
  year={2022},
  volume={},
  number={},
  pages={1-8},
  doi={10.1109/ETFA52439.2022.9921505}
}

@InProceedings{Bansal2022,
author="Bansal, Suguman",
editor="Singh, Gagandeep
and Urban, Caterina",
title="{Specification-Guided Reinforcement Learning}",
booktitle="Static Analysis",
year="2022",
publisher="Springer Nature Switzerland",
address="Cham",
pages="3--9"
}

@InProceedings{MalerN04,
author="Maler, Oded
and Nickovic, Dejan",
editor="Lakhnech, Yassine
and Yovine, Sergio",
title="Monitoring Temporal Properties of Continuous Signals",
booktitle="Formal Techniques, Modelling and Analysis of Timed and Fault-Tolerant Systems",
year="2004",
publisher="Springer Berlin Heidelberg",
address="Berlin, Heidelberg",
pages="152--166",
isbn="978-3-540-30206-3"
}

@InProceedings{DM10,
author="Donz{\'e}, Alexandre
and Maler, Oded",
editor="Chatterjee, Krishnendu
and Henzinger, Thomas A.",
title="Robust Satisfaction of Temporal Logic over Real-Valued Signals",
booktitle="Formal Modeling and Analysis of Timed Systems",
year="2010",
publisher="Springer Berlin Heidelberg",
address="Berlin, Heidelberg",
pages="92--106",
isbn="978-3-642-15297-9"
}

@article{Deshmukh2017,
author    = {Jyotirmoy V. Deshmukh and
           Alexandre Donz{\'{e}} and
           Shromona Ghosh and
           Xiaoqing Jin and
           Garvit Juniwal and
           Sanjit A. Seshia},
title     = {Robust Online Monitoring of Signal Temporal Logic},
journal   = {Formal Methods in System Design},
volume    = {51},
number    = {1},
pages     = {5--30},
year      = {2017},
}

@misc{gymnasiumLibrary,
      title={Gymnasium: A Standard Interface for Reinforcement Learning Environments}, 
      author={Mark Towers and Ariel Kwiatkowski and Jordan Terry and John U. Balis and Gianluca De Cola and Tristan Deleu and Manuel Goulão and Andreas Kallinteris and Markus Krimmel and Arjun KG and Rodrigo Perez-Vicente and Andrea Pierré and Sander Schulhoff and Jun Jet Tai and Hannah Tan and Omar G. Younis},
      year={2024},
      eprint={2407.17032},
      archivePrefix={arXiv},
      primaryClass={cs.LG},
      url={https://arxiv.org/abs/2407.17032}, 
}

@misc{Farama2023,
  author = "{The Farama Foundation - Cart Pole}",
  title = "{Cart Pole}",
  year = 2023,
  note = {2024-10-08}
}

@article{Minigrid23,
  author       = {Maxime Chevalier-Boisvert and Bolun Dai and Mark Towers and Rodrigo de Lazcano and Lucas Willems and Salem Lahlou and Suman Pal and Pablo Samuel Castro and Jordan Terry},
  title        = {Minigrid \& Miniworld: Modular \& Customizable Reinforcement Learning Environments for Goal-Oriented Tasks},
  journal      = {CoRR},
  volume       = {abs/2306.13831},
  year         = {2023},
}

@article{stable-baselines3,
  author  = {Antonin Raffin and Ashley Hill and Adam Gleave and Anssi Kanervisto and Maximilian Ernestus and Noah Dormann},
  title   = {Stable-Baselines3: Reliable Reinforcement Learning Implementations},
  journal = {Journal of Machine Learning Research},
  year    = {2021},
  volume  = {22},
  number  = {268},
  pages   = {1-8},
  url     = {http://jmlr.org/papers/v22/20-1364.html}
}

\section*{APPENDIX}

\begin{table}[h]
\caption{PPO Training Hyperparameters}
\label{table:ppo_hyperparams_minigrid}
\centering
\small

\begin{tabular}{|c||c|}
\hline
\textbf{Hyperparameter} & \textbf{Value} \\
\hline
Total Timesteps  & $5 \times 10^{5}$ \\ \hline
Learning Rate ($\alpha$) & $3 \times 10^{-4}$ \\ \hline
Rollout Steps ($n_{steps}$) & 2048 \\ \hline
Batch Size & 64 \\ \hline
Optimization Epochs ($n_{epochs}$) & 10 \\ \hline
Discount Factor ($\gamma$) & 0.99 \\ \hline
GAE Lambda ($\lambda$) & 0.95 \\ \hline
Clip Range ($\epsilon$) & 0.2 \\ \hline
Network Architecture & [256, 256] \\ \hline
\end{tabular}
\end{table}

\end{document}